\documentclass[11pt, twocolumn]{article}

\usepackage{fullpage}
\usepackage[utf8]{inputenc} 
\usepackage[T1]{fontenc}    
\usepackage{caption}
\usepackage{hyperref}       
\usepackage{url}            
\usepackage{booktabs}       
\usepackage{amsfonts}       
\usepackage{nicefrac}       
\usepackage{microtype}      
\usepackage{graphicx}       

\title{Efficient displacement convex optimization with particle gradient descent} 

\usepackage{authblk}
\author[1]{Hadi Daneshmand\thanks{hdanesh@mit.edu}}

\author[2]{Jason D. Lee}
\author[2]{Chi Jin}
\affil[1]{Laboratory for Information and Decision Systems, MIT}
\affil[2]{Department of Electrical Engineering and Computer Science, Princeton Universit}

\newcommand{\E}{\mathop{\mathbb{E}}}
\usepackage{amsmath}
\usepackage{amssymb}
\usepackage{mathtools}
\usepackage{amsthm}
\usepackage{booktabs} 
\usepackage{hyperref}

\usepackage{hyperref}
\theoremstyle{plain}
\newtheorem{theorem}{Theorem}[section]

\newtheorem{example}{Example}[section]

\newtheorem{proposition}{Proposition}[section]

\newtheorem{lemma}[theorem]{Lemma}
\newtheorem{corollary}[theorem]{Corollary}
\theoremstyle{definition}
\newtheorem{definition}[theorem]{Definition}

\theoremstyle{remark}

\usepackage{multicol}
\usepackage[textsize=tiny]{todonotes}

\newcommand{\R}{{{\mathbb {R}}}}

\begin{document}
\maketitle

\begin{abstract}
Particle gradient descent, which uses particles to represent a probability measure and performs gradient descent on particles in parallel, is widely used to optimize functions of probability measures. This paper considers particle gradient descent with a finite number of particles and establishes its theoretical guarantees to optimize functions that are \emph{displacement convex} in measures. 
Concretely, for Lipschitz displacement convex functions defined on probability over $\R^d$, we prove that $O(1/\epsilon^2)$ particles and $O(d/\epsilon^4)$ computations are sufficient to find the $\epsilon$-optimal solutions. We further provide improved complexity bounds for optimizing smooth displacement convex functions. We demonstrate the application of our results for function approximation with specific neural architectures with two-dimensional inputs.
\end{abstract}

\section{Introduction}

Optimization in the space of probability measures has wide applications across various domains, including advanced generative models in machine learning~\cite{arjovsky2017wasserstein}, the training of two-layer neural networks~\cite{chizat2018global}, variational inference using Stein's method~\cite{liu2016stein}, super-resolution in signal processing~\cite{bredies2013inverse}, and interacting particles in physics~\cite{mccann1997convexity}.    

Optimization in probability spaces goes beyond the conventional optimization in Euclidean space. \cite{ambrosio2005gradient} extends  
the notion of steepest descent in Euclidean space to the space of probability measures with the Wasserstein metric. This notion traces back to studies of the Fokker–Planck equation, a partial differential equation (PDE) describing the density evolution of Ito diffusion. The Fokker–Planck equation can be interpreted as a gradient flow in the space of probability distributions with the Wasserstein metric~\cite{jordan1998variational}. Gradient flows have become general tools to go beyond optimization in Euclidean space ~\cite{absil2009optimization,santambrogio2017euclidean,chizat2022sparse,carrillo2018measure,carrillo2022global,carrillo2021equilibrium}. 

Gradient flows enjoy a fast global convergence on an important function class called displacement convex functions~\cite{ambrosio2005gradient} which is introduced to analyze equilibrium states of physical systems~\cite{mccann1997convexity}. 
Despite their fast convergence rate, gradient flows are hard to implement.  Specifically, there are numerical solvers only for the limited class of linear functions with an entropy regularizer.

We study a different method to optimize functions of probability measures called particle gradient descent~\cite{chizat2018global,chizat2022sparse}. This method restricts optimization to sparse measures with finite support ~\cite{nitanda2017stochastic,chizat2018global,chizat2022sparse,li2022sampling} as
\begin{align}\label{eq:particle_program}
   \min_{w_1,\dots, w_n} F \left( \frac{1}{n} \sum_{i=1}^n \delta_{w_i} \right),
\end{align}
where $\delta_{w_i}$ is the Dirac measure at $w_i \in \Omega \subset \R^d$. Points $w_1, \dots, w_n$ are called particles. \textit{Particle gradient descent} 
is the standard gradient descent optimizing the particles \cite{chizat2018global,chizat2022sparse}. This method is widely used to optimize neural networks~\cite{chizat2018global}, take samples from a broad family of distributions~\cite{li2022sampling}, and simulate gradient flows in physics~\cite{carrillo2022global}. As will be discussed, $F$ is not convex in particles due to its permutation-invariance to the particles. In that regard, the convergence of particle gradient descent is not guaranteed for general functions. 

Gradient descent links to gradient flow as $n\to \infty$. In this asymptotic regime, \cite{chizat2018global} proves that the empirical distribution over the particles $w_1,\dots, w_n$ implements a (Wasserstein) gradient flow for $F$. Although the associated gradient flow globally optimizes displacement convex functions, the implication of such convergence has remained unknown for a finite number of particles.  
\subsection{Main contributions.}
We prove that particle gradient descent efficiently optimizes displacement convex functions. Consider 
 the sparse measure $\mu_n$ with support of size $n$. The error for $\mu_n$ can be decomposed as
\begin{multline*} F(\mu_n) - F^* := \\ \underbrace{F(\mu_n) -   \min_{\mu_n} F(\mu_n)}_{\text{optimization error}} + \underbrace{\min_{\mu_n} F(\mu_n)-F^*}_{\text{approximation error}}. 
\end{multline*}
The optimization error in the above equation measures how much the function value of $\mu_n$ can be reduced by particle gradient descent. The approximation error is induced by the sparsity constraint. While the optimization of particles reduces the optimization error, the approximation error is independent of the optimization and depends on $n$. 
 
\paragraph{Optimization error.}    For displacement convex functions, we establish the global convergence of variants of particle gradient descent. Table~\ref{tab:rate_summary} presents the computational complexity of particle gradient descent optimizing smooth and Lipschitz displacement convex functions.  To demonstrate the applications of these results, we provide examples of displacement convex functions that have emerged in machine learning,  tensor decomposition, and physics. 
\paragraph{Approximation error.}  Under a certain Lipschitz continuity condition, we prove the approximation error is bounded by $O(\frac{1}{\sqrt{n}})$ with a high probability. Furthermore, we prove this bound can be improved to $O(1/n)$ for convex and smooth functions in measures.  

Finally, we demonstrate the application of the established results for a specific neural network with two-dimensional inputs, and zero-one activations. When the inputs are drawn uniformly from the unit circle, we prove that $n$-neurons achieve $O(1/n)$-function approximation in polynomial time for a specific function class.
 
\begin{table*}[t!]
    \centering
    \begin{tabular}{|l|l|l|}
    \hline
    Function class & Regularity & Complexity 
    \\
     \hline
     $\lambda$-displacement convex & $\ell$-smooth  & $nd\left(\frac{\ell-\lambda}{\ell+\lambda}\right)\log(\ell/\epsilon)$ \\
    \hline
    star displacement convex & $\ell$-smooth & $nd \ell \left(\frac{1}{\epsilon}\right)$  \\
    \hline
    $\lambda$-displacement convex & $L$-Lipschitz  & $ndL^2/(\lambda \epsilon)$  \\
    \hline
    star displacement convex &$L$-Lipschitz  & $nd \ell \left(\frac{1}{\epsilon}\right)$ \\
     \hline
    \end{tabular}
    \caption{\textit{Computational complexity to reach an $\epsilon$-optimization error.} See Theorems~\ref{thm:smooth} and \ref{thm:nonsmooth} for  formal statements. }
    \label{tab:rate_summary}
\end{table*}

\section{Related works}

There are alternatives to particle gradient descent for optimization in the space of measures. For example, conditional gradient descent optimizes smooth convex functions with a sub-linear convergence rate~\cite{frank1956algorithm}. This method constructs a sparse measure with support of size $n$ using an iterative approach. This sparse measure is $O(\frac{1}{n})$-accurate in $F$~\cite{dunn1979rates,jaggi2013revisiting}.  However, each iteration of the conditional gradient method casts to a non-convex optimization without efficient solvers. Instead, the iterations of particle gradient descent are computationally efficient.

\cite{chizat2018global} establishes the link between Wasserstein gradient flows and particle gradient descent. This study proves that particle gradient descent implements the gradient flows in the limit of infinite particles for a rich function class. The neurons in single-layer neural networks can be interpreted as the particles whose density simulates a gradient flow.  The elegant connection between gradient descent and gradient flows has provided valuable insights into the optimization of neural networks~\cite{chizat2018global} and their statistical efficiency~\cite{chizat2020implicit}.  In practice, particle gradient descent is limited to a finite number of particles. Thus, it is essential to study particle gradient descent in a non-asymptotic regime. In this paper, we analyze optimization with a finite number of particles for displacement convex functions.

Displacement convexity has been used in recent studies of neural networks~\cite{javanmard,hadi22}. \cite{javanmard} establishes the global convergence of radial basis function networks using an approximate displacement convexity. \cite{hadi22} proves the global convergence of gradient descent for a single-layer network with two-dimensional inputs and zero-one loss in realizable settings. Motivated by these examples, we analyze optimization for general (non-)smooth displacement convex functions. 

Displacement convexity relates to the rich literature on geodesic convex optimization. Although 
the optimization of geodesic convex functions is extensively analyzed by ~\cite{zhang2016first,udriste2013convex,absil2009optimization}  for Riemannian manifolds, less is known for the non-Riemannian manifold of probability measures with the Wasserstein-2 metric ~\cite{jordan1998variational}.

In machine learning, various objective functions do not have any spurious local minima. This property was observed in early studies of neural networks. \cite{baldi1989neural} show that the training objective of two-layer neural networks with linear activations does not have suboptimal local minima. This proof is extended to a family of matrix factorization problems, including matrix sensing, matrix completion, and robust PCA~\cite{ge2017no}. Smooth displacement convex functions studied in this paper inherently do not admit spurious local minima~\cite{javanmard2020analysis}.

For functions with no spurious minima, escaping the saddle points is crucial, which is extensively studied for smooth functions~\cite{jin2017escape,daneshmand2018escaping}.  Although gradient descent may converge to suboptimal saddle points, random initialization effectively avoids the convergence of gradient descent to saddle points~\cite{lee2016gradient}. Yet, gradient descent may need a long time to escape  saddles~\cite{du2017gradient}. To speed up the escape, \cite{jin2017escape} leverages noise that allows escaping saddles in polynomial time. Building on these studies, we analyze the escaping of saddles for displacement convex functions.  

\section{Displacement convex functions} 

Note that the objective function $F$ is invariant to the permutation of the particles. This permutation invariance concludes that $F$ is not convex as the next Proposition states. 

\begin{proposition} \label{proposition:non-convex}
Suppose that $w_1^*, \dots, w_n^*$ is the unique minimizer of an arbitrary function $F(\frac{1}{n}\sum_{i=1}^n \delta_{w_i})$ such that $w_1^* \neq w_2^*$. If $F$ is invariant to the permutation of $w_1, \dots, w_n$, then it is non-convex. 
\end{proposition}
The condition of having distinct optimal particles, required in the last Proposition, ensures the minimizer is not a trivial minimizer for which all the particles are equal.   
Since there is no global optimization method for non-convex functions, we study the optimization of the specific family of displacement convex functions.

\subsection{Optimal transport}  To introduce displacement convexity, we need to review the basics of optimal transport theory.  
Consider two probability measures $\mu$ and $\nu$ over $\R^d$. A transport map from $\mu$ to $\nu$ is a function $T:\R^d \to \R^d$ such that 
\begin{align}
    \int_A \nu(x) dx = \int_{ T^{-1}(A)} \mu(x)dx
\end{align}
holds for any Borel subset  $A$ of $\R^d$ \cite{santambrogio2017euclidean}. The optimal transport $T^*$ has the minimum transportation cost:
\begin{align*}
    T^* = \arg\min_T \int \text{cost}(T(x), x) d\mu(x).
\end{align*}
We use the standard squared Euclidean distance function for the transportation 
cost ~\cite{santambrogio2017euclidean}. Remarkably, the transport map between distributions may not exist. For example, one can not transport a Dirac measure to a continuous measure.

In this paper, we frequently use the optimal transport map for two $n$-sparse measures in the following form \begin{equation} \label{eq:disc_measure}
 \mu = \frac{1}{n}\sum_{i=1}^n \delta_{w_i}, \quad \nu= \frac{1}{n} \sum_{i=1}^n \delta_{v_i}.\end{equation} 
For the sparse measures, a permutation of $[1,\dots, n]$, denoted by $\sigma$,   transports $\mu$ to $\nu$. Consider the set $\Lambda$, containing all permutations of $[1,\dots, n]$ and define
\begin{align}
\label{eq:optimal_permuation}
    \sigma^* = \arg\min_{\sigma \in \Lambda} \sum_{i=1}^n \| w_i - v_{\sigma(i)}\|^2.
\end{align}
The optimal permutation in the above equation yields the optimal transport map from $\mu$ to $\nu$ as $T^*(w_i) = v_{\sigma^*_i}$, and the Wasserstein-2 distance between $\mu$ and $\nu$: 
 \begin{align} \label{eq:wdist}
    W_2^2(\mu,\nu) = \sum_{i=1}^n \| w_i- v_{\sigma^*(i)}\|^2_2.
\end{align}
 Note that we omit the factor $1/n$ in $W_2^2$ for ease of notation. 
 \subsection{Displacement convex functions}  
 The displacement interpolation between $\mu$ and $\nu$ is defined by the optimal transport map  as \cite{mccann1997convexity}
\begin{align} 
    \mu_t =  \left((1-t)\text{Identity} + T^*\right)_\# \mu, 
\end{align}
where $G_\# \mu$ denotes the measure obtained by pushing  $\mu$ with $G$.
Note that the above interpolation is different from the convex combination of measures, i.e., $(1-t) \mu + t \nu$. For sparse measure, the displacement interpolation is $w_i - w_{\sigma*(i)}$ for the optimal permutation $\sigma^*$ defined in Eq.~\eqref{eq:optimal_permuation}. 

$\lambda$-displacement convexity asserts Jensen's inequality along the displacement interpolation \cite{mccann1997convexity} as 
\begin{multline*}
    F(\mu_t) \leq (1-t) F(\mu) \\ + t F(\nu) -\frac{\lambda}{2}(1-t)(t) W_2^2(\mu,\nu).
\end{multline*}
A standard example of a displacement convex function is a convex quadratic function of measures.
\begin{example}
    Consider  
\begin{align*}
   Q(\mu) = \int K(x-y) d\mu(x) d\mu(y)
\end{align*}
where $\mu$ is a measure over $\R^d$ and $K(\Delta)$ is convex in $\Delta \in \R^d$; then, $Q$ is $0$-displacement convex~\cite{mccann1997convexity}. 
\end{example}
The optimization of $Q$ over a sparse measure is convex~\footnote{$Q$ does not satisfy the condition of Proposition~\ref{proposition:non-convex}.}. However, this is a very specific example of displacement convex functions. Generally, displacement convex functions are not necessarily convex.

Recall the sparse measures defined in Eq.~\eqref{eq:disc_measure}. While convexity asserts Jensen's inequality for the interpolation of $\{w_i\}$ with all $n!$ permutations of $\{v_j\}$, displacement convexity only relies on a specific permutation. In that regard, displacement convexity is weaker than convexity. In the following example, we elaborate on this difference. 
\begin{example} \label{example:energy}
The energy distance between measures over $\R$ is defined as
\begin{multline} \label{eq:mmd}
    E(\mu,\nu) = 2 \int | x - y | d\mu(x) d\nu(y) \\- \int | x - y | d\mu(x) d\mu(x) - \int| x- y | d\nu(x) d\nu(y).
\end{multline}
$E(\mu,\nu)$ is $0$-displacement convex in $\mu$~\cite{carrillo2018measure}.
\end{example}
 According to Proposition~\ref{proposition:non-convex}, $E$ does not obey Jensen's inequality for interpolations with an arbitrary transport map. In contrast, $E$ obeys Jensen's inequality for the optimal transport map, since it is monotone in $\R$~\cite{carrillo2018measure}. This key property concludes $E$ is displacement convex.     

Remarkably, the optimization of the energy distance has applications in machine learning and physics.
\cite{hadi22} show that the training of two-layer neural networks with two-dimensional inputs (uniformly drawn from the unit sphere) casts to minimizing $E(\mu,\nu)$ in a sparse measure $\mu$. The optimization of the energy distance has been also used in clustering \cite{szekely2017energy}. In physics, the gradient flow on the energy distance describes interacting particles from two different species \cite{carrillo2018measure}.

\subsection{Star displacement convex functions} 
Our convergence analysis extends to a broader family of functions.  Let $\widehat{\mu}$ denote the optimal $n$-sparse solution for the optimization in Eq.~\eqref{eq:particle_program}, and $\mu_t$ is obtained by the displacement interpolation between $\mu$ and $\widehat{\mu}$. Star displacement convex function $F$ obeys
\begin{align*}  
      \sum_{i} \langle w_i - T(w_i), \partial_{w_i} F(\mu) \rangle  \geq F(\mu) - F(\widehat{\mu}),
\end{align*}
where $T$ is the optimal transport map from $\mu$ to $\widehat{\mu}$. The above definition is inspired by the notion of star-convexity~\cite{nesterov2006cubic}.  It is easy to check that $0$-displacement convex functions are star displacement convex. 

Star displacement convex optimization is used for generative models in machine learning.
An important family of generative models optimizes the Wasserstein-2 metric~\cite{arjovsky2017wasserstein}. Although Wasserstein 2 is not displacement convex \cite{santambrogio2017euclidean}, it is star displacement convex.    
\begin{example} \label{lemma:weakconvex_w2}
$W_2^2(\mu,\nu)$ is star displacement convex in $\mu$ as long as $\mu$ and $\nu$ has sparse supports of the same size. 
\end{example}
Star displacement convexity holds for complete orthogonal tensor decomposition. Specifically, we consider the following example of tensor decomposition.
\begin{example} \label{example:tensor}
Consider the orthogonal complete tensor decomposition of order $3$, namely 
\begin{align*} 
  \min_{w_1,\dots, w_d \in \R^d} \left( G\left(\frac{1}{n}\sum_{i=1}^d \delta_{w_i}\right)=- \sum_{i=1}^d \sum_{j=1}^d \left\langle \frac{w_j}{\|w_j\|}, v_i\right\rangle^3\right),
\end{align*}
where $v_1, \dots, v_d$ are orthogonal vectors over the unit sphere denoted by $\mathcal{S}_{d-1}$. 
\end{example}
 Although orthogonal tensor decomposition is not convex~\cite{anandkumar2014tensor}, the next lemma proves that it is star displacement convex. 
\begin{lemma} \label{theorem:star_tensor}
$G$ is star displacement convex for $w_1, \dots, w_n \in \mathcal{S}_{d-1}$. 
\end{lemma}
To prove the above lemma, we leverage the properties of the optimal transport map used for displacement interpolation.

There are more examples of displacement convex functions in machine learning~\cite{javanmard2020analysis} and physics~\cite{carrillo2009example}. Motivated by these examples, we analyze displacement convex optimization.

\section{Optimization of smooth functions}
Gradient descent is a powerful method to optimize smooth functions that enjoy a dimension-free convergence rate to a critical point~\cite{nesterov1999global}. More interestingly, a variant of gradient descent converges to local optimum~\cite{daneshmand2018escaping,jin2017escape,ge2015escaping,xu2018first,zhang2017hitting}. Here, we prove gradient descent globally optimizes the class of (star) displacement convex functions. Our results are established for the standard gradient descent, namely the following iterates
\begin{multline}
    w_{i}^{(k+1)} = w_i^{(k)} - \gamma \partial_{w_i} F(\mu_k),\\  \mu_k := \frac{1}{n}\sum_{i=1}^n \delta_{w_i^{(k)}}
\end{multline} 
where $\partial_{w_i} F$ denotes the gradient of $F$ with respect to $w_i$.
The next Theorem establishes the convergence of gradient descent. 
\begin{theorem}\label{thm:smooth}
Assume $F$ is $\ell$-smooth, and particle gradient descent starts from distinct particles $w_1^{(0)}\neq \dots \neq w_n^{(0)}$. Let $\widehat{\mu}$ denote the optimal solution of \eqref{eq:particle_program}. 
\begin{itemize}
    \item[(a)] For $(\lambda>0)$-displacement functions,
\begin{multline*}
    F(\mu_{k+1}) - F(\widehat{\mu}) \\\leq \ell  \left(1-\left(\frac{2 \lambda \ell \gamma}{\ell+\lambda}\right)\right)^k  W_2^2(\mu_0,\widehat{\mu})
\end{multline*}
holds as long as $\gamma\leq 2/(\lambda+\ell)$.

\item[(b)] Under $0$-displacement convexity,  
\begin{multline*}
      F(\mu_{k+1}) - F(\widehat{\mu}) \\ \leq \frac{2 (F(\mu_0) - F(\widehat{\mu}))W_2^2(\mu_0,\widehat{\mu})}{2 W_2^2(\mu_0,\widehat{\mu}) + (F(\mu_0) - F(\widehat{\mu})) \gamma k }
 \end{multline*}
 holds for $\gamma \leq 1/\ell$.
 \item[(c)] Suppose that $F$ is star displacement convex and $\max_{m\in\{1,\dots, k\}} W_2^2(\mu_m,\widehat{\mu})\leq r^2$; then 
 \begin{multline*}
      F(\mu_{k+1}) - F(\widehat{\mu}) \\\leq \frac{2 (F(\mu_0) - F(\widehat{\mu}))r^2}{2 r^2 + (F(\mu_0) - F(\widehat{\mu})) \gamma k }
 \end{multline*}
 holds for $\gamma \leq 1/\ell$.
\end{itemize}
\end{theorem}

\begin{table*}[t!]  
    \centering
    \begin{tabular}{|l|l|}
    \hline
    Function class & Convergence rate
    \\
     \hline
     $\lambda$-disp. convex  & $\ell  \left(\frac{\ell-\lambda}{\ell+\lambda}\right)^k  W_2^2(\mu_0,\widehat{\mu})$ \\
    
    $\lambda$-strongly convex & $\frac{\ell}{2}  \left(\frac{\ell-\lambda}{\ell+\lambda}\right)^k  \sum_{i}\| w_i - w^*_i\|^2_2$  \\
    \hline
    $0$-disp. convex  & $2L W_2^2(\mu_0,\widehat{\mu})k^{-1}$  \\
  
      convex  & $2L \left(\sum_{i}\| w_i - w^*_i\|^2_2\right)(k+4)^{-1}$ \\
     \hline
    \end{tabular}
   
    \caption{\textit{Convergence rates for the optimization of $\ell$-smooth functions.} We use the optimal choice for the stepsize $\gamma$ to achieve the best possible rate. Recall $\widehat{\mu} = \frac{1}{n}\sum_{i=1}^n\delta_{w_i^*}$ denotes the optimal solution for Eq.~\eqref{eq:particle_program}. Rates for convex functions:   \cite{nesterov1999global}. Rates for displacement convex functions: Theorem~\ref{thm:smooth}.   }
\label{tab:ratesmooth}
\end{table*}

Table~\ref{tab:ratesmooth} compares convergence rates for convex and displacement convex functions. We observe an analogy between the rates. The main difference is the replacement of Euclidean distance by the Wasserstein distance in the rates for displacement convex functions. This replacement is due to the permutation invariance of $F$.
The Euclidean distance between $(w_1^*, \dots, w_n^*)$ and permuted particles $(w_{\sigma(1)}^*, \dots , w_{\sigma(n)}^*)$ can be arbitrary large, while $F$ is invariant to the permutation of particles $w_1,\dots, w_n$. Proven by the last Theorem~\ref{thm:smooth}, Wasserstein distance effectively replaces the Euclidean distance for permutation invariant displacement convex functions.

Smooth displacement convex functions are non-convex, hence have saddle points. However, displacement convex functions do not have suboptimal local minima \cite{javanmard2020analysis}. Such property has been frequently observed for various objective functions in machine learning. To optimize functions without suboptimal local minima, escaping saddle points is crucial since saddle points may avoid the convergence of gradient descent~\cite{ge2015escaping}. \cite{lee2016gradient} proves that random initialization effectively avoids the convergence of gradient descent to saddle points. Similarly, the established global convergence results rely on a weak condition on initialization:  The particles have to be distinct. A regular random initialization satisfies this condition.

 Escaping saddles with random initialization may require considerable time for general functions. \cite{du2017gradient} propose a smooth function on which escaping saddles may take an exponential time with the dimension. Notably, the result of the last theorem holds specifically for displacement convex functions. For this function class, random initialization not only enables escaping saddles but also leads to global convergence.

\section{Optimization of Lipschitz functions}
Various objective functions are not smooth. For example,  the training loss of neural networks with the standard ReLU activation is not smooth. In physics, energy functions often are not smooth~\cite{mccann1997convexity,carrillo2022global}. Furthermore, recent sampling methods are developed based on non-smooth optimization with particle gradient descent~\cite{li2022sampling}. Motivated by these broad applications, we study the optimization of non-smooth displacement convex functions. In particular, we focus on $L$-Lipschitz functions whose gradient is bounded by $L$. 


To optimize non-smooth functions, we add noise to gradient iterations as
\begin{multline}
\tag{PGD}\label{eq:noisy_descent}
    w_{i}^{(k+1)} = w_i^{(k)} \\- \gamma_k \left( \partial_{w_i} F(\mu_k) + \frac{1}{\sqrt{n}}\xi_i^{(k)} \right) 
\end{multline}
where $\xi_1^{(k)}, \dots \xi_n^{(k)} \in \R^d$ are random vectors uniformly drawn from the unit ball. The above perturbed gradient descent (PGD) is widely used in smooth optimization to escape saddle points~\cite{ge2015escaping}. The next Theorem proves this random perturbation can be leveraged for optimization of non-smooth functions, which are (star) displacement convex. 

\begin{theorem}\label{thm:nonsmooth}
Consider the optimization of a $L$-Lipschitz function  with ~\ref{eq:noisy_descent} starting from $w_1^{(0)}\neq \dots \neq w_n^{(0)}$. 
\begin{itemize}
    \item[a.] If $F$ is $\lambda$-displacement convex, then
\begin{align*}
\min_{k \in \{1, \dots, m \}} \left\{ \E \left[ F(\mu_k) - F(\widehat{\mu})\right] \right\} & \leq \frac{2(L^2 +1)}{\lambda(m+1)}
\end{align*}
holds for $\gamma_k = 2/(\lambda(k+1))$.
\item[b.] If $F$ is star displacement convex, then 
\begin{multline*}
\min_{k \in \{1, \dots, m \}} \left\{ \E \left[ F(\mu_k) - F(\widehat{\mu}) \right] \right\} \\\leq \frac{1}{\sqrt{m}}\left( W_2^2(\mu_0,\widehat{\mu}) + L +1\right)
\end{multline*}
holds for $\gamma_1 = \dots = \gamma_m = 1/\sqrt{m}$.
\end{itemize}
Notably, the above expectations are taken over random vectors $\xi_1^{(k)}, \dots \xi_n^{(k)}$.
\end{theorem}

Thus, \ref{eq:noisy_descent} yields an $\epsilon$-optimization error with $O(1/\epsilon^2)$ iterations to reach $\epsilon$-suboptimal solution for Lipschitz displacement convex functions. This rate holds for the optimization of the energy distance since it is $2$-Lipschitz and $0$-displacement convex~\cite{carrillo2018measure}.  
 \cite{hadi22} also establishes the convergence of gradient descent on the specific example of the energy distance. The last Theorem extends this convergence to the general function class of non-smooth Lipschitz displacement convex functions. While the convergence of \cite{hadi22} is in the Wasserstein distance, our convergence results are in terms of the function value.

 \section{Approximation error} \label{sec:app_error}
 Now, we turn our focus to the approximation error. We provide bounds on the approximation error for two important function classes: 
\begin{itemize}
    \item[(i)] Lipschitz functions in measures. 
    \item[(ii)] Convex and smooth functions in measures. 
\end{itemize}

 For (i), we provide the probabilistic bound $O\left(\frac{1}{\sqrt{n}}\right)$ on the approximation error; then, we improve the bound to $O(\frac{1}{n})$ for (ii).

\subsection{Lipschitz functions in measures} We introduce a specific notion of Lipschitz continuity for functions of probability measures. This notion relies on Maximum Mean Discrepancy (MMD) between probability measures. Given a positive definite kernel $K$, MMD$_K$ is defined as  
\begin{multline*}
    \left(\text{MMD}_K(\mu, \nu)\right)^2 = \int K(w,v) d\mu(w) d\mu(v) \\- 2 \int K(w,v) d\mu(w) d\nu (v) + \int K(w,v) d\nu(w) d\nu(v)
\end{multline*}
\noindent
MMD is widely used for the two-sample test in machine learning~\cite{gretton2012kernel}.
Leveraging MMD, we define the following Lipschitz property.  

\begin{definition}[$L$-MMD$_K$ Lipschitz]
$F$ is $L$-MMD$_K$ Lipschitz, if there exists a positive definite Kernel $K$ such that 
\begin{align*}
    | F(\mu) - F(\nu)| \leq L \times \text{MMD}_K(\mu,\nu)
\end{align*}
holds for all probability measures $\mu$ and $\nu$.
\end{definition}
Indeed, the above Lipschitz continuity is an extension of the standard Lipschitz continuity to functions of probability measures. A wide range of objective functions obeys the above Lipschitz continuity. Particularly, \cite{chizat2018global} 
introduces a unified formulation for training two-layer neural networks, sparse deconvolution, and tensor decomposition as 
\begin{align} \label{eq:nn_chizat}
     R\left(\int \Phi(w) d\mu(w) \right) 
\end{align}
where $\Phi: \R^d \to \mathcal{H}$ is a map whose range lies in the Hilbert space $\mathcal{H}$ and $R: \mathcal{H} \to \R_+$. Under a weak assumption, $R$ is $L$-MMD$_K$ Lipschitz. 

\begin{proposition} \label{prop:Lipschitz}
 If $R$ is $L$-Lipschitz in its input, then it is $L$-MMD$_K$-Lipschitz for $K(w,v) = \langle \Phi(w),\Phi(v) \rangle$.
\end{proposition}

Thus, the class of Lipschitz functions is rich. For this function class, $O(\frac{1}{\sqrt{n}})$-approximation error is achievable.

\begin{proposition} 
\label{lemma:approximation} Suppose that there exists a uniformly bounded kernel $\|K\|_{\infty} \leq B$ such that $F$ is $L$-MMD$_K$ Lipschitz; then,
\begin{align*}
\min_{\mu_n} F(\mu_n)-F^* \leq  \frac{3\sqrt{B}}{\sqrt{n}} 
\end{align*}
holds with probability at least $1- \exp(-1/n)$. 
\end{proposition}
The last Proposition is a straightforward application of Theorem 7 in \cite{gretton2012kernel}.
Combining the above result with  Theorem~\ref{thm:nonsmooth} concludes the total complexity of $O(d/\epsilon^4)$ to find an $\epsilon$-optimal solution for Lipschitz displacement functions. The complexity can be improved to $O(d/\epsilon^2)$ for smooth functions according to Theorem~\ref{thm:smooth}.

The established bound $O(1/\sqrt{n})$ can be improved under assumptions on the kernel $K$ associated with the Lipschitz continuity. For $d$-differentiable shift-invariant kernels, \cite{xu2022accurate} establishes a considerably tighter bound $O(\frac{\log(n)^d}{n})$ when the support of the optimal measure is a subset of the unit hypercube.

\subsection{Convex functions in measures}
If $F$ is convex and smooth in $\mu$, we can get a tighter bound on the approximation error.
\begin{lemma} \label{lemma:approx_convex}
Suppose $F$ is convex and smooth in $\mu$. If the probability measure $\mu$ is defined over a compact set, then 
\begin{align*}
    \min_{\mu_n} F(\mu_n) - F^* = O\left(\frac{1}{n}\right)
\end{align*}
holds for all $n$.
\end{lemma}
The proof of the last Lemma is based on the convergence rate of the Frank-Wolfe algorithm~\cite{jaggi2013revisiting}. This algorithm optimizes a smooth convex function by adding particles one by one. After $n$ iterates, the algorithm obtains an $n$-sparse measure which is $O(1/n)$-suboptimal. \cite{bach2017breaking} uses this proof technique to bound the approximation error for neural networks. The last lemma extends this result to a broader function class.  

Remarkably, the energy distance is convex and smooth in $\mu$, hence enjoys $O(1/n)$-approximation error as stated in the next lemma. 
\begin{lemma} \label{lemma:energy_distance}
    $E(\mu,\nu)$ is convex and smooth in $\mu$ when $\mu$ and $\nu$ have a bounded support.
\end{lemma}

\subsection{Applications for neural networks}
The established theoretical analysis has a subtle application for the function approximation with neural networks. Consider the class of functions in the following form 
\begin{align}
    f(x) = \int \varphi(x^\top w) d\nu(w)
\end{align}
where $x, w \in \R^2$ lies on the unit circle and $\nu$ is a measure with support contained in the upper-half unit circle. $\varphi$ is the standard zero-one ridge function:
\begin{align}
    \varphi(a) = \begin{cases} 
    1 & a >0 \\ 
    0 & a \leq 0
    \end{cases}.
\end{align}
The above function is used in the original MacCulloch-Pitts model for neural networks~\cite{mcculloch1943logical}. To approximate function $f$, one may use a neural network with a finite number of neurons implementing the following output function: 
\begin{align}
    f_n(x) = \frac{1
    }{n} \sum_{i=1}^n \varphi(x^\top w_i),
\end{align}
where $w_1, \dots, w_n$ are points over the unit circle representing the parameters of the neurons. To optimize the location of $w_1,\dots, w_n$,
one may minimize the standard mean-squares loss as 
\begin{align}
    \min_{w_1,\dots,w_n} \left( L(w):= \E_x  \left( f_n(x) - f(x)\right)^2 \right).
\end{align}

As is stated in the next corollary, \ref{eq:noisy_descent} optimizes $L$ up to the approximation error when the input $x$ is distributed uniformly over the unit circle.
\begin{corollary} \label{cor:neural_nets}
Suppose that the input $x$ is drawn uniformly over the unit circle. After a specific transformation of the coordinates for $w_1,\dots, w_n$, \ref{eq:noisy_descent} with $n$ particles with stepsize $\gamma_k =1/\sqrt{k}$ obtains $w^{(k)}:=[w_1^{(k)},\dots, w_n^{(k)}]$ after $k$ iteration such that 
\begin{align}
    \E \left[ L(w^{(k)})\right] =O(\frac{n}{\sqrt{k}}+\frac{1}{n})
\end{align}
holds where the expectation is taken over the algorithmic randomness of \ref{eq:noisy_descent}.
\end{corollary}
The last corollary is the consequence of part b of Theorem~\ref{thm:nonsmooth}, and the approximation error established in Lemma~\ref{lemma:approx_convex}. For the proof, we use the connection between $L$ and the energy distance derived by \cite{hadi22}. While \cite{hadi22} focuses on realizable settings, the last corollary holds for non-realizable settings when the measure $\nu$ is not an $n$-sparse measure.
\section{Experiments} 
We experimentally validate established bounds on the approximation and optimization error. Specifically, we validate the results for the example of the energy distance, which obeys the required conditions for our theoretical results.  
\subsection{Optimization of the energy distance}
\begin{figure}
     \centering
     \includegraphics[width=0.4\textwidth]{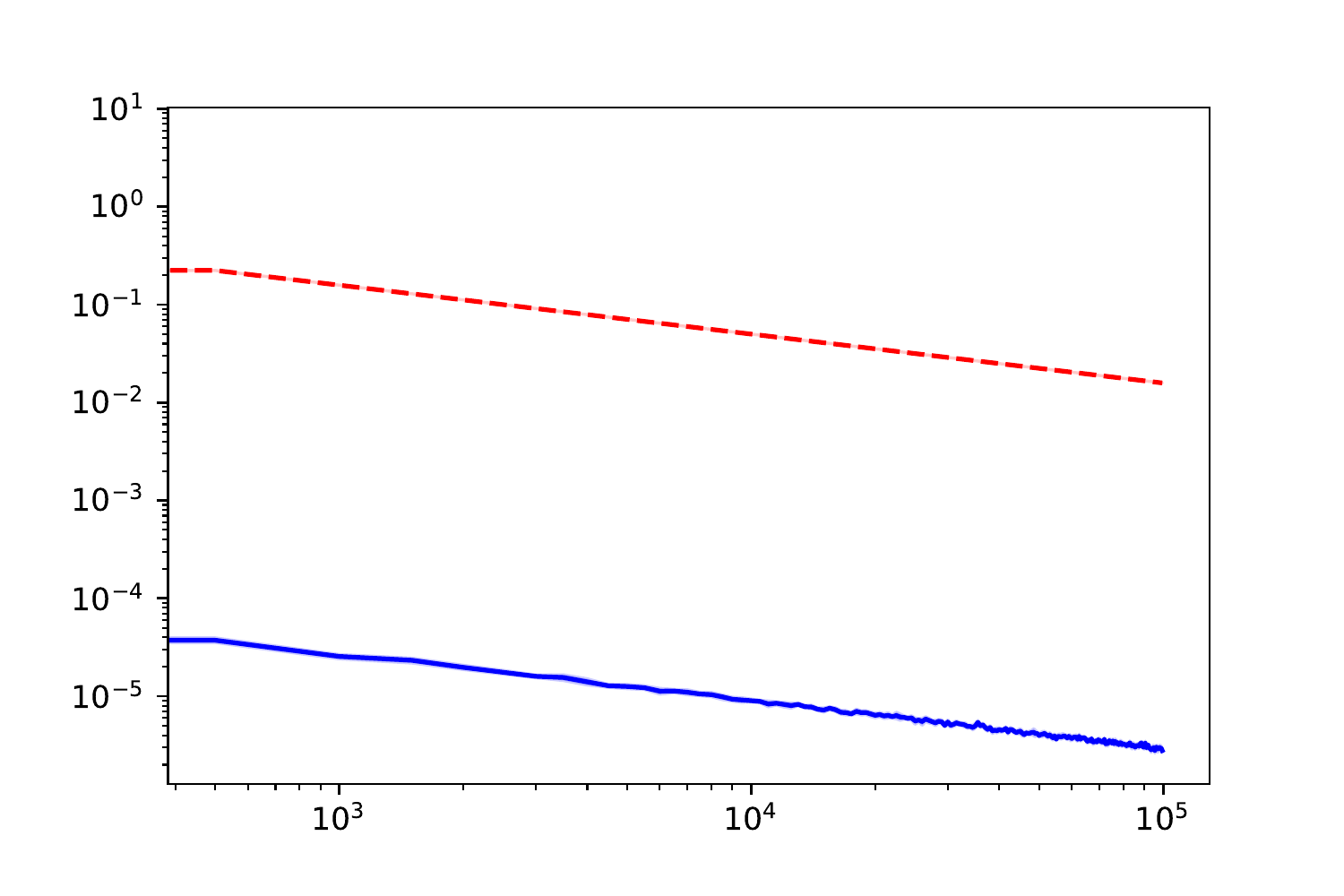}
     \caption{\footnotesize{\textbf{The convergence of \ref{eq:noisy_descent} for the energy distance.} Horizontal: $\log(k)$; vertical: $\log(E(\mu_k,\nu)- E(\widehat{\mu},\nu))$. The red dashed line is the theoretical convergence rate. The blue line is the convergence observed in practice for the average of 10 independent runs. }}
     \label{fig:energy_distance}
 \end{figure}

As noted in Example~\ref{example:energy}, the energy distance is displacement convex. Furthermore, it is easy to check that this function is $2$-Lipschitz. For the sparse measures in Eq.~\eqref{eq:disc_measure}, the energy distance has the following form 
\begin{multline*}
    n^2 E(\mu,\nu) = 2\sum_{i,j=1}^n |w_i - v_j| \\
    - \sum_{i,j=1}^n | v_i - v_j | - \sum_{i,j=1}^n | w_i - w_j|,  
\end{multline*}
where $n=100$ for this experiment.
We draw $v_1,\dots, v_n$ at random from uniform$[0,1]$. Since $E$ is not a smooth function, we use \ref{eq:noisy_descent} to optimize $w_1,\dots, w_n \in \R$. In particular, we use $\xi_1^{(k)}$ i.i.d. from uniform$[-0.05,0.05]$. For the stepsize, we use $\gamma_k = 1/\sqrt{k}$ required for the convergence result in Theorem~\ref{thm:nonsmooth} (part b). In  Figure~\ref{fig:energy_distance}, we observe a match between the theoretical and experimental convergence 
 rate for \ref{eq:noisy_descent}. 
 
\subsection{Approximation error for the energy distance
}
 Lemma~\ref{lemma:approx_convex} establish $O(1/n)$ approximation error for convex functions of measures. Although the energy distance $E(\mu,\nu)$ is not convex in the support of $\mu$, it is convex and smooth in $\mu$ as stated in Lemma~\ref{lemma:energy_distance}. Thus, $O(1/n)$-approximation error holds for the energy distance. We experimentally validate this result. Consider the recover of $\nu=$uniform$[-1,1]$ by minimizing the energy distance as 
 \begin{multline*}
     E(\mu,\nu) = \frac{2}{n}\sum_{i=1}^n | w_i - v| d\nu(v) \\ - \frac{1}{n^2}\sum_{i,j=1}^n | w_i- w_j|  - \int | v- v'|d\nu(v) d\nu(v').
 \end{multline*}
The above integrals can be computed in closed forms using $\int_{-1}^1|w-v|d\nu(v) = w^2+1$. Hence, we can compute the derivative of $E$ with respect to $w_i$. We run \ref{eq:noisy_descent} with stepsize determined in the part b of Theorem~\ref{thm:nonsmooth} for $k=3\times10^5$ iterations and various $n \in \{2^2, \dots, 2^8\}$. Figure~\ref{fig:energy_distance} shows how the error decreases with $n$ in the log-log scale. In this plot, we observe that $E$ enjoys a mildly better approximation error compared to the established bound $O(1/n)$. 
\begin{figure}
    \centering
    \includegraphics[width=0.4\textwidth]{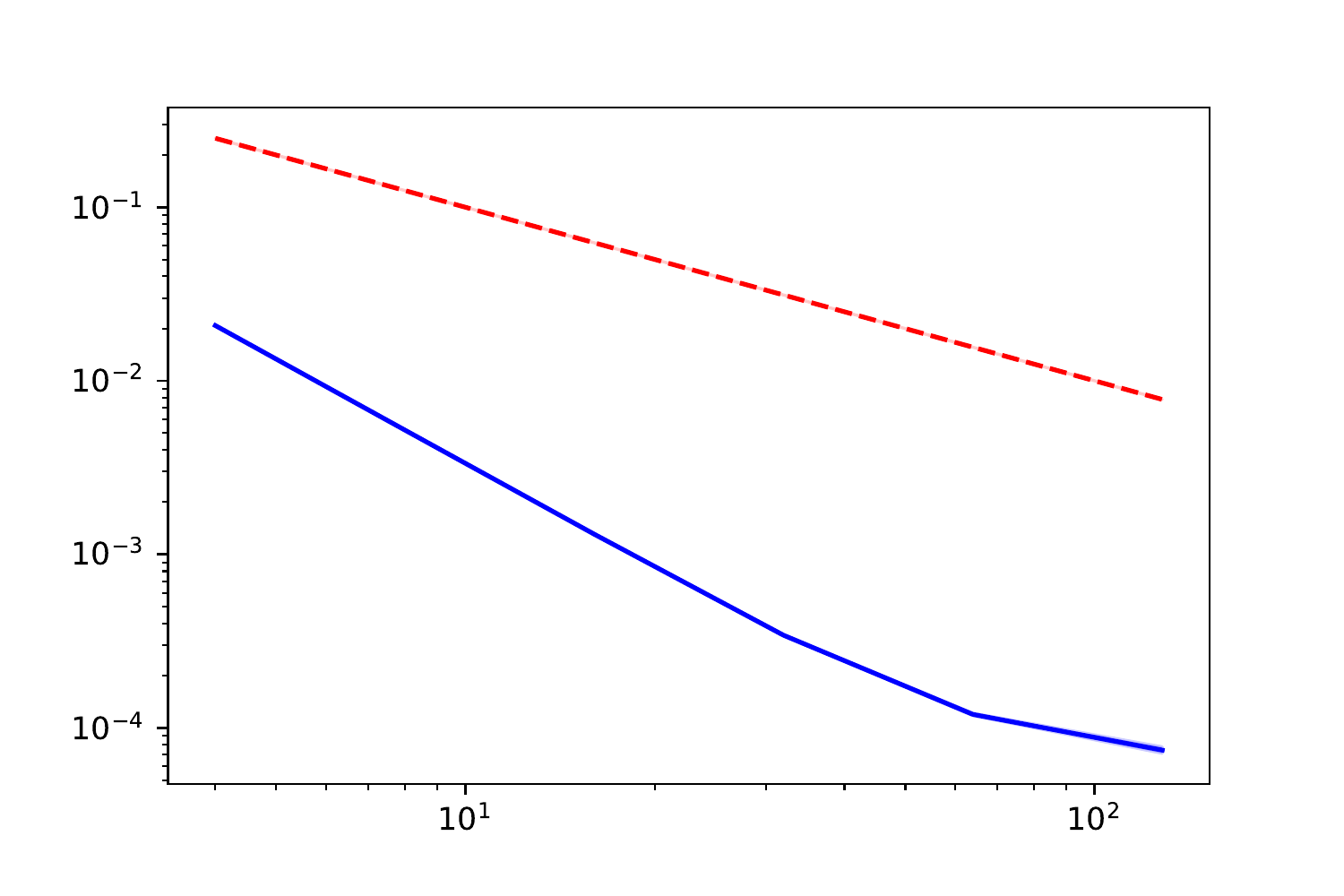}
    \caption{\footnotesize{\textbf{Approximation error for the energy distance.} Horizontal: $n$; vertical: $E(\widehat{\mu}_n, \nu)$ where $\widehat{\mu}_n$ is obtained by $3\times 10^5$ iterations of \ref{eq:noisy_descent} with $n$ particles. The red dashed line is the theoretical $O(1/n)$-bound for the approximation error. The plot is in the $\log$-scale for both axes. The (blue) plot shows the average of 10 independent runs. }}
    \label{fig:my_label}
\end{figure}
\section{Discussions}
We establish a non-asymptotic convergence rate for particle gradient descent when optimizing displacement convex functions of measures. Leveraging this convergence rate, we prove the optimization of displacement convex functions of (infinite-dimensional) measures can be solved in polynomial time with input dimension, and the desired accuracy rate. This finding will be of interest to various communities, including the communities of non-convex optimization, optimal transport theory, particle-based sampling, and theoretical physics. 

The established convergence rates are limited to particle gradient descent. Yet, there may be other algorithms that converge faster than this algorithm. Convex optimization literature has established lower-bound complexities required to optimize convex function (with first-order derivatives)~\cite{nesterov1999global}. Given that displacement convex functions do not obey the conventional notion of convexity, it is not clear whether these lower bounds extend to this specific class of non-convex functions. More research is needed to establish (Oracle-based) lower-computational-complexities for displacement convex optimization.

 Nesterov's accelerated gradient descent enjoys a considerably faster convergence compared to gradient descent in convex optimization. Indeed, this method attains the optimal convergence rate using only first-order derivatives of smooth convex functions~\cite{nesterov1999global}. This motivates future research to analyze the convergence of accelerated gradient descent on displacement convex functions.

We provided examples of displacement convex functions, including the energy distance. Displacement convex functions are not limited to these examples. A progression of this work is to assess the displacement convexity of various non-convex functions. In particular, non-convex functions invariant to permutation of the coordinates, including latent variable models and matrix factorization~\cite{anandkumar2014tensor}, may obey displacement convexity under weak assumptions.

A major limitation of our result is excluding displacement convex functions with entropy regularizers that have emerged frequently in physics~\cite{mccann1997convexity}. The entropy is displacement convex. Restricting the support of measures to a sparse set avoids the estimation of the entropy. Thus, particle gradient descent is not practical for the optimization of functions with the entropy regularizer. To optimize such functions, the existing literature uses a system of interacting particles solving a stochastic differential equation~\cite{philipowski2007interacting}. In asymptotic regimes, this algorithm implements a gradient flow converging to the global optimal measure~\cite{philipowski2007interacting}. To assess the complexity of these particle-based algorithms, we need non-asymptotic analyses for a finite number of particles. 
\section*{Acknowledgments and Disclosure of Funding}
We thank Francis Bach, Lenaic Chizat and Philippe Rigollet for their helpful discussions on the related literature on particle-based sampling, the energy distance minimization and Riemannian optimization. This project was funded by the Swiss National Science Foundation (grant P2BSP3\_195698).

\bibliographystyle{unsrt}  
\bibliography{refs}

\end{document}